%% file: root.tex
\documentclass[letterpaper, 10 pt, conference]{ieeeconf}  

\IEEEoverridecommandlockouts                              

\usepackage[normalem]{ulem}
\usepackage[utf8]{inputenc}
\usepackage[T1]{fontenc}
\usepackage[textsize=scriptsize]{todonotes}
\usepackage{lipsum}
\usepackage{algorithm}
\usepackage{algpseudocode}
\usepackage{url}
\usepackage{multirow}
\usepackage{blindtext}
\usepackage[separate-uncertainty = true, free-standing-units, space-before-unit, use-xspace]{siunitx}
\usepackage{color,soul}
\usepackage{svg}
\usepackage{amsmath} 
\usepackage{amsfonts}
\usepackage{amssymb}  

\newcommand{\bds}[1]{\boldsymbol{#1}}

\input{review.tex} 

\title{\LARGE \bf

Learning to Adapt the Parameters of Behavior Trees and Motion Generators (BTMGs) to Task Variations 
}

\author{Faseeh Ahmad$^{1}$, Matthias Mayr$^{1}$, and Volker Krueger$^{1}$
	\thanks{*This work was partially supported by the Wallenberg AI, Autonomous Systems and Software Program (WASP) funded by Knut and Alice Wallenberg Foundation.}
	\thanks{$^{1}$Department of Computer Science, Faculty of Engineering (LTH), Lund University, SE~221~00 Lund, Sweden. E-mail: <firstname>.<lastname>@cs.lth.se.
	}%
}

\begin{document}

\maketitle
\thispagestyle{empty}
\pagestyle{empty}

\begin{abstract}
The ability to learn new tasks and quickly adapt to different variations or dimensions is an important attribute in agile robotics.
In our previous work, we have explored Behavior Trees and Motion Generators (BTMGs) as a robot arm policy representation to facilitate the learning and execution of assembly tasks. 
The current implementation of the BTMGs for a specific task may not be robust to the changes in the environment and  may not generalize well to different variations of tasks.
We propose to extend the BTMG policy representation with a module that predicts BTMG parameters for a new task variation. 
To achieve this, we propose a model that combines a Gaussian process and a weighted support vector machine classifier. This model predicts the performance measure and the feasibility of the predicted policy with BTMG parameters and task variations as inputs. Using the outputs of the model, we then construct a surrogate reward function that is utilized within an optimizer to maximize the performance of a task over BTMG parameters for a fixed task variation.
To demonstrate the effectiveness of our proposed approach, we conducted experimental evaluations on push and obstacle avoidance tasks in simulation and with a real \textit{KUKA iiwa} robot. Furthermore, we compared the performance of our approach with four baseline methods.
\end{abstract}

\section{Introduction}

Robots have been utilized effectively for many years in repetitive and automated industrial processes. However, despite the shift towards smaller batch sizes and increased demand for customization, many robot systems still require a lengthy and expensive reconfiguration process. To keep up with the demands of society and modern industrial production, robots should have the ability to adapt quickly to different situations. 
In these situations, the task formulations should be robust to failures, interpretable, and possibly reactive to failures. Additionally, the task formulations should also be adaptable to different variations or dimensions of the same task, such as pushing an object to different locations, picking an object from any location in the space, and avoiding an obstacle with different shapes and positions.

To overcome the challenges, Rovida F. et al.~\cite{rovida18btmg} have suggested a representation that combines behavior trees (BT)~\cite{colledanchise142iicirsa, colledanchise17bt} and motion generators (MG), (BTMG). 
In our previous work, we used BTMGs to model skills for contact-rich tasks such as inserting a peg into the hole to mimic engine assembly~\cite{rovida18btmg, mayr21iros} and pushing an object to a target location~\cite{mayr2022combining,mayr2022skill}.

\begin{figure}
    \centering
    \includegraphics[width=\columnwidth]{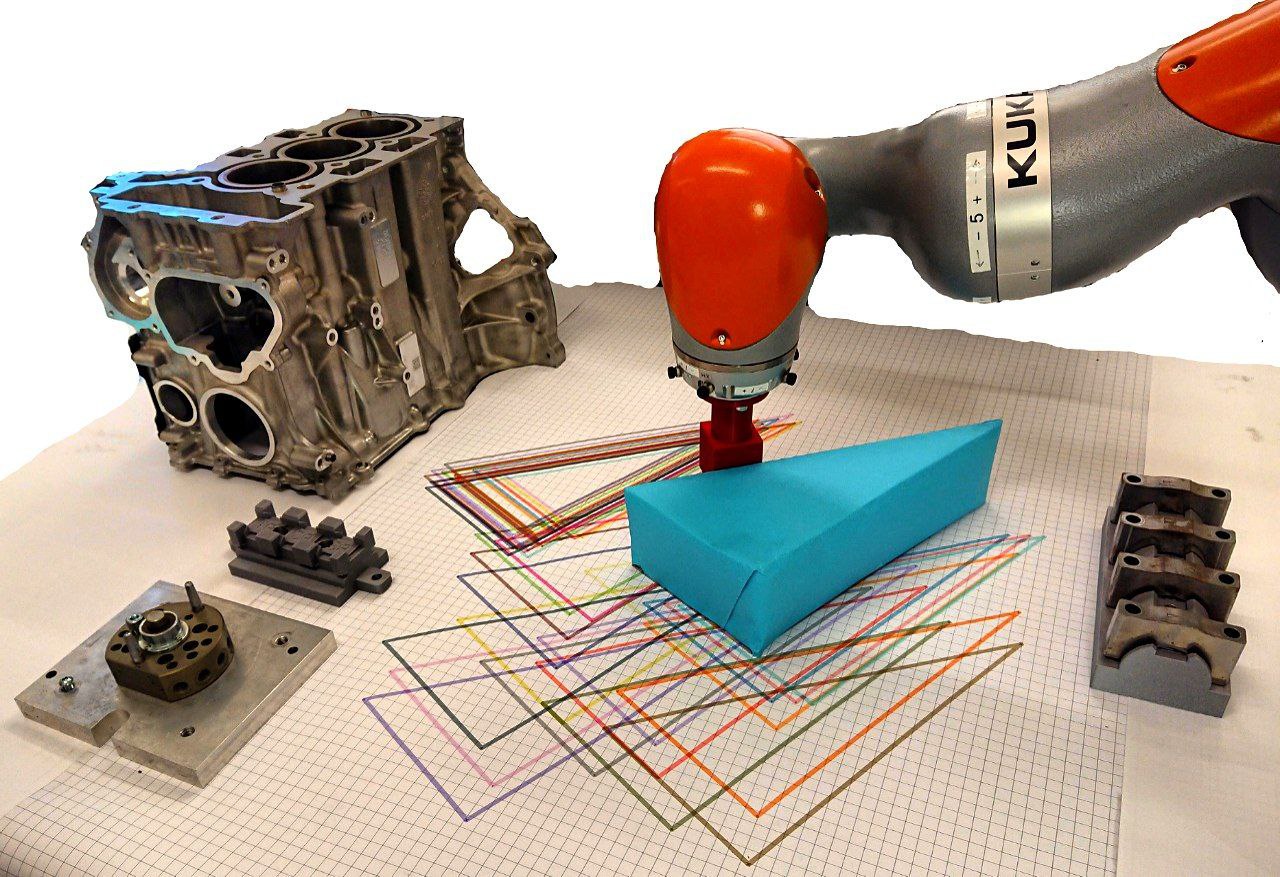}
    \caption{The experimental setup. It shows the object with the skewed weight distribution that is pushed with a \SI{45}{\milli\meter} wide peg. On the table the different start and goal positions for the object can be seen in different colours. On the sides, some example sizes for obstacles are shown.}
    \label{fig:exp-setup}
\end{figure}

A BTMG is a parameterized policy representation that combines the strengths of both behavior trees and motion generators. Behavior trees provide a clear and intuitive way to describe the high-level logic of the robot's behavior, while motion generators generate the low-level motion commands by controlling the end-effector in Cartesian space. For a more concrete definition of motion generators, refer to ~\cite{rovida18btmg}. The parameters of a BTMG can be used to specify the structure of the behavior tree as well as values such as controller stiffness. 

BTMGs are easy to interpret and can be designed to be \textbf{robust} to faults and failures that can occur during execution~\cite{rovida18btmg}. Furthermore, they have the ability to be \textbf{reactive}~\cite{colledanchise142iicirsa}, allowing the robot to adapt and respond to current circumstances.
Simple BTs can also be systematically combined with more complex ones to solve complex tasks ~\cite{rovida18btmg, mayr21iros, rovida172iicirsi}.

BTMGs are a promising technique for motion modeling because of their explicitness, robustness, and reactiveness.
There are mainly three ways to set the parameters of BTMGs. One way is to specify them manually or fine-tune them by experts~\cite{rovida18btmg}. Another way is to determine those parameters through reasoning. However this requires the existence of such a reasoner for the task at hand, which can not always be assumed. Finally, BTMG parameters can be learned through reinforcement learning (RL)~\cite{mayr2022combining, mayr2022skill, mayr22priors}. However, learned BTMG parameters are in many cases scenario-specific and changes in the setup may require relearning them. 

Setting BTMG parameters using these methods can limit the usage of BTMGs in scenarios that require quick adaptability. 
For example, tasks such as pushing an object to different locations, picking an object from various locations, or even picking objects with various shapes would require updating the parameters of the respective BTMGs. 
This problem is also present in the original formalization of dynamic motion primitives (DMPs)~\cite{976259,ijspeert2002learning} and was later addressed in~\cite{ijspeert2013dynamical}.

In this paper, we propose an extension to the BTMG formulation that enables quick adaptation to different task variations by incorporating a model that combines a Gaussian process (GP) and a weighted support vector machine (SVM) classifier. 
Our model uses a GP to learn a function that predicts the performance measure of a policy using task variations and BTMG parameters as inputs. Furthermore, the model also trains a weighted SVM classifier that predicts the feasibility of a  policy. 
For example, in a push task, the performance measure of a policy can be given by its overall reward, which depends on the error between the actual and target position of the pushed object. In this task, a policy can be feasible when this error is below a user-defined threshold. Once the model is trained, we optimize the BTMG parameters over the resulting surrogate reward function for a given new task variation.

The following are our main contributions:
\begin{itemize}
    \item We extend BTMG policy representation that enables it to quickly adapt to task variations.
    \item We propose a model that combines a GP and a weighted SVM classifier to predict the performance measure and feasibility of a BTMG policy for a new task variation, and subsequently optimize the output of the model to obtain resulting BTMG parameters.
    \item We evaluate the performance of the proposed method in simulation and on a real \textit{KUKA iiwa} robot for two tasks and compare its performance with four baselines.
\end{itemize}
\section{Related Work}
Movement primitives, based on motor primitives theory~\cite{mussa1999modular,flash2005motor}, are mathematical formulations of dynamic systems that generate motions.  Two well-known movement primitives used in robotics are Dynamic Movement Primitives (DMPs)~\cite{976259,ijspeert2002learning} and Probabilistic Movement Primitives (ProMPs)\cite{paraschos2013probabilistic}.  Movement primitives can be generalized and have proven successful in various robotics applications,  such as dynamic motion primitives ~\cite{976259,ijspeert2002learning}. Similar to our BTMGs, DMPs intially lacked the capacity to generalize to different task parameters. This was resolved later by introducing a small change in the transformation system~\cite{ijspeert2013dynamical}.

While both DMPs and BTMGs are capable of generating motions through attractor landscapes, the parameters for DMPs are learned implicitly from a set of demonstrations, whereas parameters for BTMGs can be explicitly specified manually, inferred through a reasoner, or learned using RL. Nevertheless, a comprehensive comparison of the two approaches would require further investigation and is outside the scope of this paper.

DMPs have been extended with intermediate via points~\cite{ning2011accurate,ning2012novel, weitschat2018safe, zhou2019learning}, and can generalize to new goals by interpolating weights of neighboring DMPs~\cite{weitschat2013dynamic} or by using Gaussian Process Regression (GPR) to generate new parameters~\cite{forte11ras}. Furthermore, GPs~\cite{williams2006gaussian}  have been used to generalize DMPs to external task variations, arbitrary movements, and adapting trajectories to new situations online in~\cite{alizadeh2016learning, fanger2016gaussian, forte11ras}, respectively. In~\cite{lee18joss}, Gaussian mixture models are used to learn the mapping of task parameters and the forcing term of DMPs. 

The mixture of movement primitives (MoMP) algorithm introduced in~\cite{muelling2010learning, mulling2013learning}, can also be used to generalize the basis movements stored in the library. The MoMP algorithm captures the robot's position and velocity as parameters for the expected hitting position and velocity. A new motion is generated by a weighted sum of DMPs, assigning a probability to a DMP based on the sensed state. MoMPs and ProMPs have been applied successfully in various applications, including learning striking movements for table tennis robots~\cite{Muelling_ICHR_2012, Gomez-Gonzalez_PICHR_2016} and solving Human-Robot collaborative tasks~\cite{gjm_2016_AURO_c} using ProMPs.

We draw inspiration from prior work on DMPs to extend BTMG's formulation by incorporating generalization to different task variations using GP, as seen in~\cite{forte11ras, alizadeh2016learning, fanger2016gaussian}.  These studies employed GPs to directly map task variations to DMP parameters, which we refer to as the \textit{direct} model in this paper. However, our approach differs significantly in how we use GPs. Instead of using the \textit{direct} model, we propose a model that combines GP with a weighted SVM classifier to predict the performance of tasks and the feasibility of a policy, using task variations and BTMG parameters as inputs. Since our model predicts both performance measure and feasibility, we refer to it as the \textit{PerF} model, short for performance and feasibility.
\section{BTMG and Task Variations}
We define BTMG as a parametric policy representation, $\text{BTMG}(\bds{\theta})$ where $\bds{\theta} \in \mathbb{R}^N$. The parameters $\bds{\theta}$ can range from determining the structure of the behavior tree (BT) to specifying the controller stiffness values of the motion generator (MG). These parameters are further subdivided into \textbf{intrinsic} parameters $\bds{\theta}_i$ and \textbf{extrinsic} parameters $\bds{\theta}_e$~\cite{ahmad2022generalizing}.

Intrinsic parameters $\bds{\theta}_i$ determine the structure of the behavior tree, the number of control nodes, the type of motion generator, etc. 
For example, consider a policy $T_p$ for a push task, which has intrinsic parameters $\bds{\theta}_i$. These parameters are fixed and independent of the task instance, meaning that $T_p$ uses the same $\bds{\theta}_i$ values regardless of the starting position, or the target position of the object. In other words, $\bds{\theta}_i$ is situation-invariant. Within the scope of this paper, these parameters are assumed to be known a priori.

Extrinsic parameters $\bds{\theta}_e$ are situation dependent e.g. to determine the applied force, offsets, and the velocity of the end effector. Again, $\bds{\theta}_e$ can be specified manually~\cite{rovida18btmg, rovida16icaps}, inferred through a reasoning framework, or learned using RL. We have already demonstrated how RL can be used to obtain BTMG parameters~\cite{mayr21iros} and used it in simulation and on a real robot to solve multi-objective tasks~\cite{mayr2022skill, mayr22priors}.

In addition to $\bds{\theta}$, we also consider task variations $\bds{v} \in \mathbb{R}^M$. 
Task variations refer to different possible alterations of a given task, such as different start and goal positions of an object. For example, a task variation $\bds{v}$ in the case of a push task would be a $4\text{D}$ vector consisting of the values of the start and goal positions of the object along the horizontal and vertical axes. 

\begin{figure}[tpb!]
	{
        \vspace{0.1cm}
		\setlength{\fboxrule}{0pt}
		\framebox{\parbox{3in}{
  		\includegraphics[width=1\columnwidth]{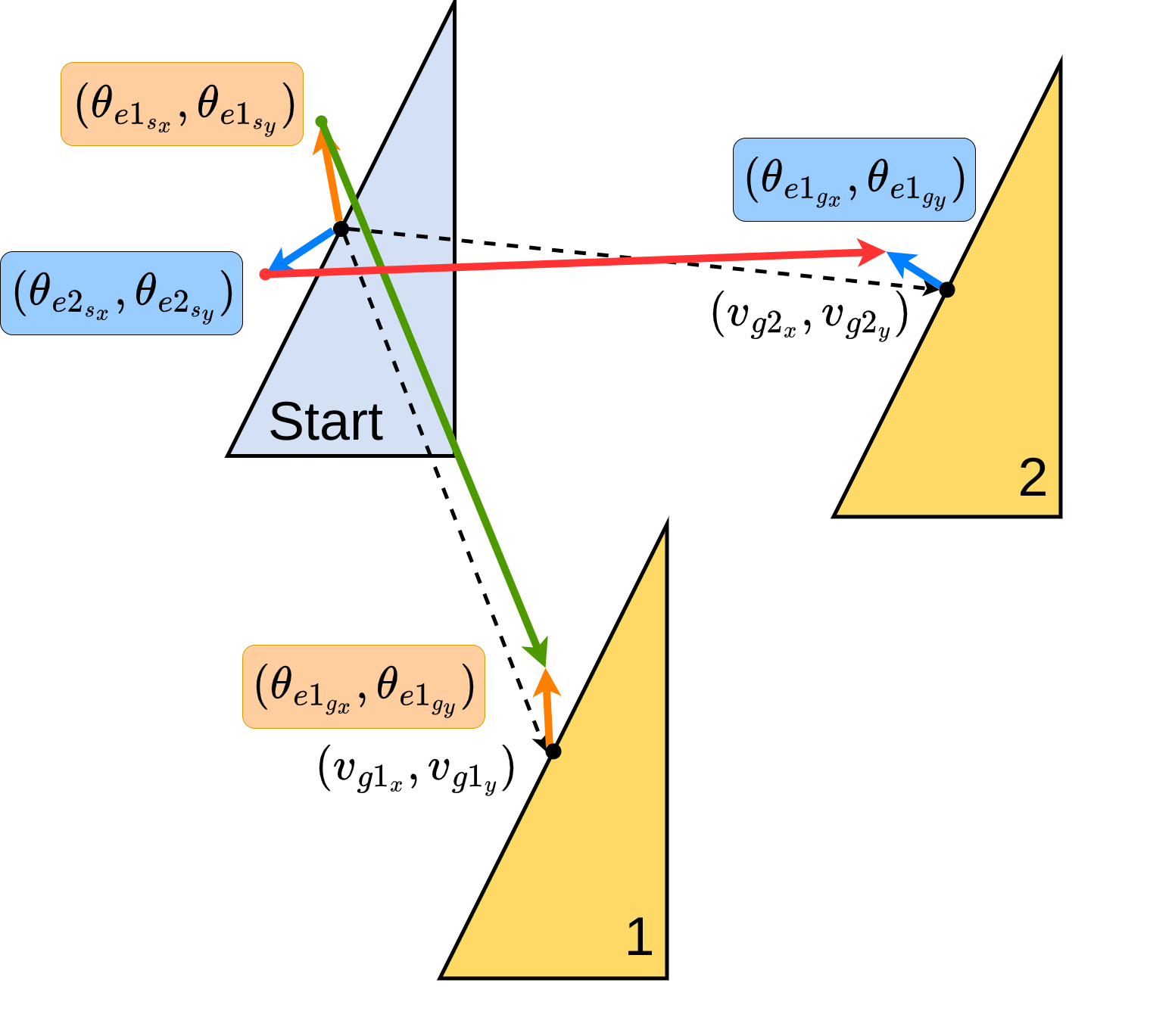}
        }
	}
 }
	\caption{An illustration of two simplified task variations $\bds{v}_1$ and $\bds{v}_2$ in the pushing task that only vary the goal location. The orange and blue vectors are set by the respective learned extrinsic parameters $\bds{\theta}_{e1}$ and $\bds{\theta}_{e2}$, so that they define the resulting green and red push vectors that should successfully push the object. } 
	\label{fig:push-parameters}
\end{figure}

Note that the task variation parameters are different from the extrinsic BTMG parameters (Figure~\ref{fig:push-parameters}). We take two task variations $\bds{v}_1 = (v_{s_{x}}, v_{s_{y}}, v_{g1_{x}}, v_{g1_{y})}$ and $\bds{v}_2= (v_{s_{x}}, v_{s_{y}}, v_{g2_{x}}, v_{g2_{y})}$ that define the start and goal positions of the object. For variations $\bds{v}_1$ and $\bds{v}_2$, we have corresponding $\bds{\theta}_{e1} = (\theta_{e1_{s_{x}}}, \theta_{e1_{s_{y}}}, \theta_{e1_{g_{x}}}, \theta_{e1_{g_{y}}})$ and $\bds{\theta}_{e2} = (\theta_{e2_{s_{x}}}, \theta_{e2_{s_{y}}}, \theta_{e2_{g_{x}}}, \theta_{e2_{g_{y}}})$ that collectively define the start and the goal locations for the pushing action.

As $\bds{\theta}_i$ has no impact on adapting BTMGs to different variations, our objective in this paper is to establish a relationship between $\bds{\theta}_e$ and $\bds{v}$ that would enable the adaptation of BTMGs to new variations.
\section{Approach}
\label{sec:approach}
In this section, we explain how we adapt BTMG parameters for a new task variation by using the \textit{PerF} model.
Figure~\ref{fig:flowchart} shows how the \textit{PerF} model works in comparison with a \textit{direct} model. The overall approach is divided into the training (Sec.~\ref{sec:trainingphase}) and query phase (Sec.~\ref{sec:Query Phase}). In the training phase, we pass each task variation $\bds{v}_k \in \mathbb{V}_{train}$, into an extended RL pipeline similar to~\cite{mayr2022skill}.
For each learning process for different task variations, we utilize three sets of outputs from the RL pipeline to train the \textit{direct} and the \textit{PerF} models:
\begin{enumerate}
\item \textit{Best policies}: For every task variation we get the best performing policy:\\
$\mathbb{T}=\{(\bds{v}_k, \bds{\theta}^*_{e,\bds{v}_k})|k=1, \ldots, n\}$

\item \textit{All evaluated configurations and their rewards:}\\
$\mathbb{K}= \{(\bds{v}_k, \bds{\theta}_{ei,\bds{v}_k}, r_{\bds{\theta}_{ei,\bds{v}_k}})|k=1, \ldots, n \text{ and } i=1, \ldots, t \leq t_\text{max}\}$

\item \textit{All evaluated configurations and their feasibility:}\\
$\mathbb{E}= \{(\bds{v}_k, \bds{\theta}_{ei,\bds{v}_k}, f_{\bds{\theta}_{ei,\bds{v}_k}})|k=1, \ldots, n \text{ and } i=1, \ldots, t \leq t_\text{max}\}$
\end{enumerate}

The \textit{direct} model $M$ is trained with the set $\mathbb{T}$ and, as a result, learns to predict $\hat{\bds{\theta}}_e$ given $\bds{v}$. On the other hand, the \textit{PerF} model is trained with the sets $\mathbb{K}$ and $\mathbb{E}$ and as a result it learns to predict the reward $\hat{r}$ and feasibility $\hat{f}$ of a policy with parameters $\bds{\theta}_e$. The model further uses $\hat{r}$ and $\hat{f}$ to generate a surrogate reward function that obtains $\hat{\bds{\theta}}_e$ given $\bds{v}$. For more details on how we obtain set $\mathbb{T}$, we direct the reader to~\cite{mayr2022skill}. To obtain sets $\mathbb{K}$ and $\mathbb{E}$, we follow the same procedure as in~\cite{mayr2022skill}, retaining all configurations along with their respective rewards and feasibilities for a given task variation.

The intuition behind using the \textit{PerF} model together with an optimizer is to guide the combination of GP and weighted SVM towards predicting policy parameters $\bds{\theta}_e$ that prioritize performance measure and feasibility. In contrast, the \textit{direct} model does not take into account the performance measure and feasibility. In the following subsections, we explain our approach in more depth.
\subsection{Training Phase}
\label{sec:trainingphase}
\begin{figure}[tpb]
	{
        \vspace{0.1cm}
		\setlength{\fboxrule}{0pt}
            \begin{center}
                
		\framebox{\parbox{3in}{
  		\includegraphics[width=0.9\columnwidth]{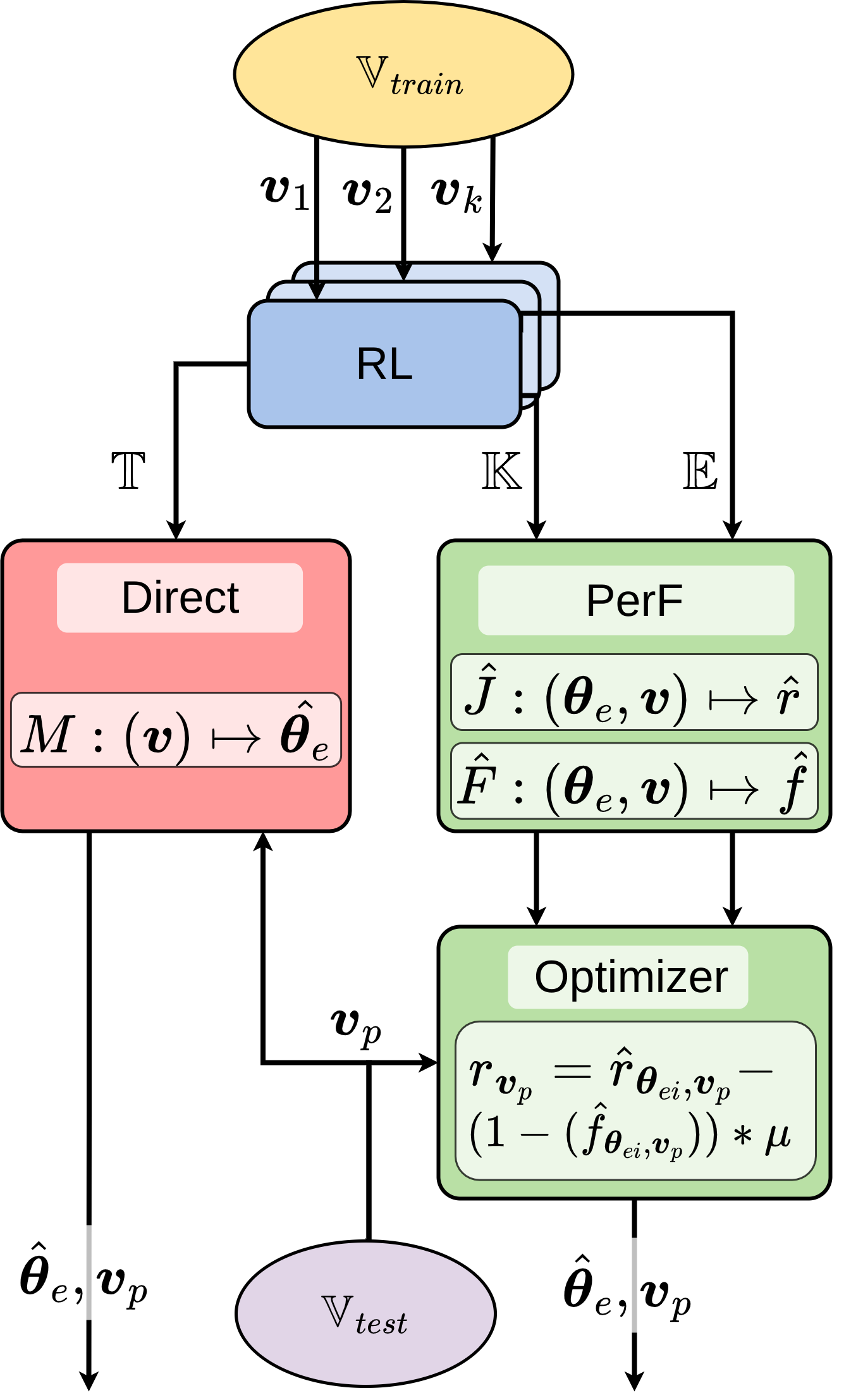}
            }
            }
            \end{center}
        }
	\caption{The pipeline of our approach and the \textit{direct} model baseline. For every task variation $\mathbf{v}$, an RL problem is solved and the respective results are provided to the GP models. When querying for a new task variation $\mathbf{v}_{p}$ both models are queried for a set of extrinsic parameters $\bds{\hat{\theta}}_{e}$.}
	\label{fig:flowchart}
\end{figure}

We frame the mapping of the task variations $\bds{v}$ to the extrinsic BTMG parameters $\bds{\theta}_e$ as a supervised learning problem. The training phase aims to learn two functions: $\hat{J}$ that predicts the reward achieved by a policy and $\hat{F}$ that predicts if a policy is feasible, see Figure~\ref{fig:flowchart}.  We propose to use GP and weighted SVM to learn $\hat{J}:(\bds{\theta}_e,\bds{v})\mapsto\hat{r}\in\mathbb{R}$ and $\hat{F}:(\bds{\theta_e},\bds{v})\mapsto\hat{f}\in\{0,1\}$. $\hat{J}$ and $\hat{F}$ are trained by data points in sets $\mathbb{K}$ and $\mathbb{E}$, provided by the RL pipeline introduced in \cite{mayr21iros}. 

For each task variation, $\bds{v}_k \in \mathbb{V}_{train}$, similar to ~\cite{mayr21iros,mayr2022skill}, we define $J_{\bds{v}_k}(\bds{\theta}_{e})$ as the expected sum of individual rewards over time, given a sequence of extrinsic parameters $\bds{\theta}_{e1}, \bds{\theta}_{e2}, \ldots, \bds{\theta}_{et} \in \bds{\theta}_e$. 

In ~\cite{mayr21iros,mayr2022skill}, we use Bayesian optimization (BO) as a black-box optimization method to obtain the optimal policy parameters $\bds{\theta}^*_{e}$ and the best reward $J_{\bds{v}_k}(\bds{\theta}^*_{e})$. In this paper, however, we use BO to obtain $J_{\bds{v}_k}(\bds{\theta}_{e})$ by computing $J_{\bds{v}_k}(\bds{\theta}_{e1}), J_{\bds{v}_k} (\bds{\theta}_{e2}), \ldots, J_{\bds{v}_k}(\bds{\theta}_{et})$ over the sequence $\bds{\theta}_{e1}, \bds{\theta}_{e2}, \ldots, \bds{\theta}_{et}$. This allows us to not only have the optimal policy parameters $\bds{\theta}^*_{e}$ and the corresponding best reward $J_{\bds{v}_k}(\bds{\theta}^*_{e})$ but it also provides us with intermediate $\bds{\theta}_{et}$ and $J_{\bds{v}_k}(\bds{\theta}_{et})$. Overall, this provides us with large amount of data to train the $\hat{J}$ function and allows us to capture the overall reward landscape better. 

In addition to learning the reward function $\hat{J}$, we also learn the feasibility function $\hat{F}$. The motivation behind learning $\hat{F}$ is twofolds: First, it provides a user-defined metric to evaluate the feasibility of a policy and second, it complements the reward formulation of a task by addressing the potential shortcomings of inaccurate reward formulations. In principle, we do not need to optimize feasibility if the reward formulation covers all aspects of the task. However, in practice, reward formulation is challenging, so feasibility addresses these shortcomings effectively. It ensures learned policies align with the task's requirements, despite imperfect reward formulations.
 
For a given task variation $\bds{v}_k$, we define the feasibility function $F_{\bds{v}_k}(\bds{\theta}_{e})$ 
as a binary function that maps to 1 or 0 depending on whether the policy achieves a user-defined metric of feasibility or not. Similar to $J_{\bds{v}_k}(\bds{\theta}_{e})$, we obtain $F_{\bds{v}_k}(\bds{\theta}_{e})$ by computing $F_{\bds{v}_k}(\bds{\theta}_{e1}), F_{\bds{v}_k}(\bds{\theta}_{e2}), \ldots, F_{\bds{v}_k}(\bds{\theta}_{et})$ for the sequence of evaluations $\bds{\theta}_{e1}, \bds{\theta}_{e2}, \ldots, \bds{\theta}_{et}$. 
For more details about the pipeline, we refer the reader to the policy optimization section in~\cite{mayr21iros,mayr2022skill}. 

To model $\hat{J}$ and $\hat{F}$, we obtain a sequence of BTMG parameter vectors, ${\bds{\theta}_{e1}, \bds{\theta}_{e2}, \ldots, \bds{\theta}_{et}}$, along with their corresponding reward values $J_{\bds{v}_k}(\bds{\theta}_{e1}), J_{\bds{v}_k}(\bds{\theta}_{e2}), \ldots, J_{\bds{v}_k}(\bds{\theta}_{et})$ and feasibility values $F_{\bds{v}_k}(\bds{\theta}_{e1}), F_{\bds{v}_k}(\bds{\theta}_{e2}), \ldots, F_{\bds{v}_k}(\bds{\theta}_{et})$ for task variations. We then use these data points to train a GP and a weighted SVM classifier. This enables us to effectively model the underlying $J$ and $F$.
\subsection{Query Phase}
\label{sec:Query Phase}
The goal of this phase is to query the trained model with a new task variation $\bds{v}_p \in \mathbb{V}_{test}$ and obtain a $\hat{\bds{\theta}}_e$ by optimizing $\hat{J}(\bds{\theta}_{et}|\bds{v}_p)$ under the feasibility constraint $\hat{F}(\bds{\theta}_{et}|{\bds{v}_p})$ (Figure~\ref{fig:flowchart}). For this purpose, we use the $\hat{J}$ and $\hat{F}$ obtained in the training phase. We solve this as an optimization problem over a sequence of $\bds{\theta}_{e}$ for a new $\bds{v}_p$.

We begin the optimization process by specifying the optimizer type, the bounds for $\bds{\theta}_e$, and the maximum number of iterations $t_\text{max}$. In our experiments, we used the Limited-memory Broyden–Fletcher–Goldfarb–Shanno (L-BFGS)~\cite{byrd1995limited,zhu1997algorithm} algorithm, which refines an initial estimate of $\bds{\theta}_{e1}$ to iteratively obtain improved evaluation points $\bds{\theta}_{et}$, where $t \leq t_\text{max}$, using the derivative as the driving function. For each new task variation $\bds{v}_k$, we run the optimizer to obtain a sequence of evaluation points $\bds{\theta}_{et}$.

Using $\hat{J}$ and $\hat{F}$, we define a surrogate reward \mbox{$r_{\bds{v}_{p}} = \hat{r}_{\bds{\theta}_{et},\bds{v}_{p}} - (1-\hat{f}_{\bds{\theta}_{et},\bds{v}_{p}})) * \mu$}. Here, the first term corresponds to the output reward value computed by $\hat{J}$, while the second term penalizes the reward if $\hat{f}_{\bds{\theta}_{et},\bds{v}_{p}}$ maps to 0. We penalize the reward $\hat{r}_{\bds{\theta}_{ei},\bds{v}_{p}}$ by a small factor $\mu$. 
We query the surrogate reward $r_{\bds{v}_{p}}$ for defined number of iterations or until the optimizer converges. 

After the optimization phase, we select the $\bds{\theta}_{et}$ that maximizes both $\hat{J}(\bds{\theta}_{et}|\bds{v}_p)$ and is feasible $\hat{F}(\bds{\theta}_{et}|\bds{v}_p)$. 
\section{Experiments}
\label{sec:experiments}
We evaluated the efficacy of our approach in simulation and also by transferring of the simulation results to a real \textit{KUKA iiwa} manipulator for two tasks: an obstacle avoidance task and a pushing task, each having its own challenges. For simulation, we utilized the \textit{DART} simulation toolkit \cite{lee18joss} and in both simulation and reality, the robot arm was controlled using a Cartesian impedance controller~\cite{mayr2022c++}, which helps reduce the disparities between simulation  and reality. Additionally, for the push task, we further reduce the sim-to-real gap by adjusting the friction coefficient appropriately. For more detailed information on bridging the sim-to-real gap, please refer to~\cite{mayr21iros}.

To train our model, we considered 20 task variations that are learned for the same amount of iterations each. Using the method detailed in Sec.~\ref{sec:trainingphase}, we train the GP and the weighted SVM classifier with the resulting BTMG parameters, the feasibility, and the reward values. The weights of the SVM classifier are adjusted automatically to adjust bias induced by an unequal number of feasible and non-feasible policies. We then tested our approach on 20 unknown task variations. This experiment is repeated five times for both tasks to show the robustness of the approach. 

We compare the performance of our approach with four baselines: 
\begin{figure}[tpb]
	{
        \vspace{0.1cm}
		\setlength{\fboxrule}{0pt}
            \begin{center}
		\framebox{\parbox{3in}{
		      \includegraphics[width=0.9\columnwidth]{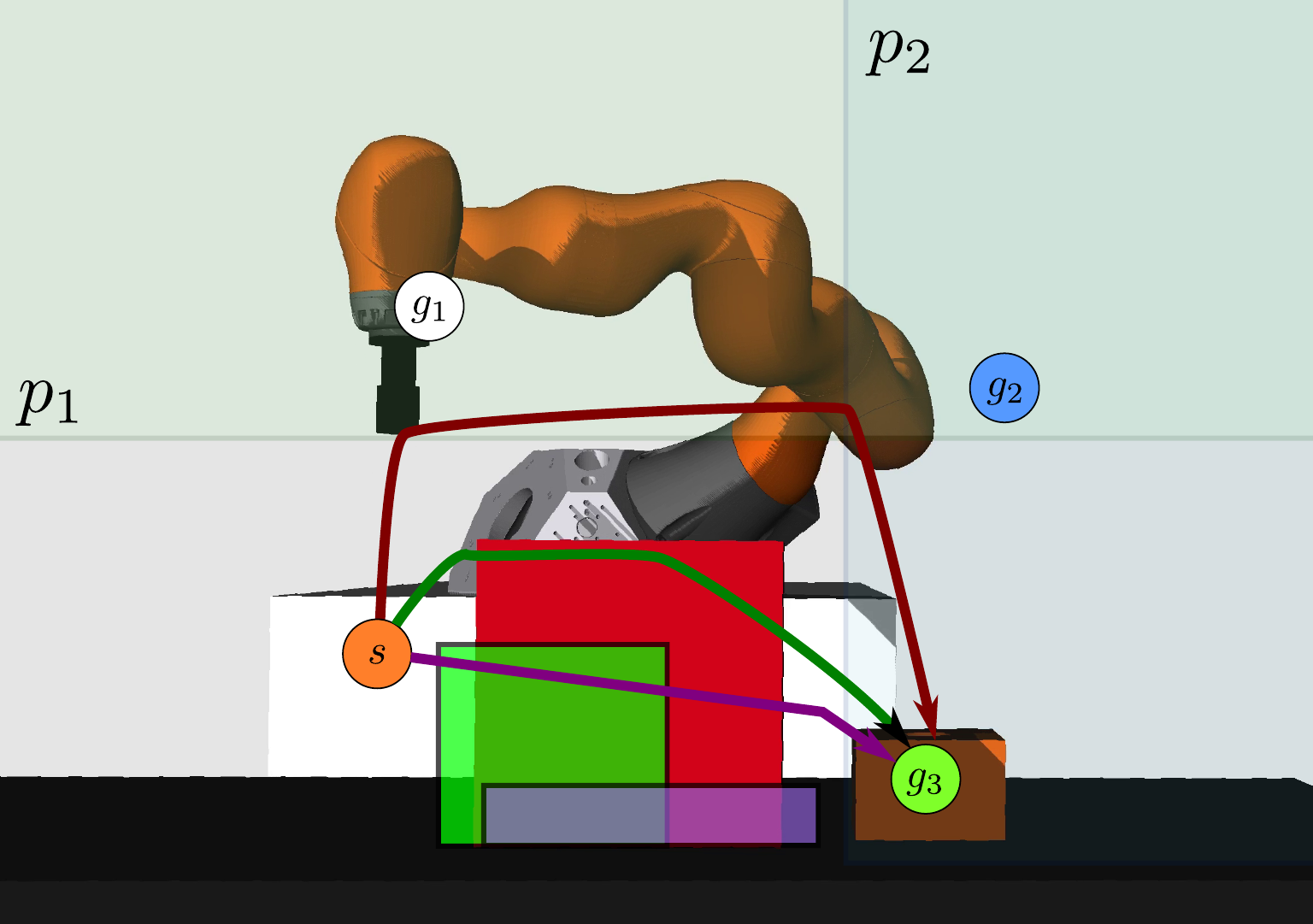}
            }
	       }
            \end{center}
    }
	\caption{The obstacle task with some of the variations of the object location, width, and height. For each object configuration, valid example trajectories are shown in the same color. For the red trajectory, the intermediate goal points ($\mathbf{g}_1$ and $\mathbf{g}_2$) and two motion switching thresholds ($p_1$ and $p_2$) are shown.}
	\label{fig:obstacle-task}
\end{figure}

\begin{enumerate}
    \item \textit{Learned}: This baseline uses the RL pipeline described in~\cite{mayr2022skill} to learn the BTMG parameters directly for the test variations. It shows which performance could be achieved if a new variation is learned from scratch instead of querying the model. Notably, our training data is generated in this way.
    \item \textit{Direct}: This model takes the best parameters for the training variations ($\mathbb{T}$) and learns a direct mapping from task variations to BTMG parameters without explicitly learning the reward.
    \item \textit{Nearest Neighbor}: For each test variation, we select the closest task variation in the training set and choose the corresponding BTMG parameters.
    \item \textit{Single Policy}: The learned BTMG parameters of a single training variation are used for all test variations. This baseline shows how well and how often the learned parameters for one task variation can be utilized in a different one without any changes.
\end{enumerate}
Although our baselines may seem simplistic, they are deliberately selected to provide insights into the functionality and performance of our approach.
Each of these baselines serves a specific purpose in understanding the capabilities and limitations of our approach.

We consider task-specific reward functions for both tasks. The rewards and feasibility measures for the tasks are defined separately in their respective sections. 

\subsection{Obstacle Avoidance Task}
\label{sec:exp-obstacle}
The objective of the obstacle avoidance task is to move the robot's end effector from the start to the goal location while avoiding an obstacle in the workspace. As shown in Fig.~\ref{fig:obstacle-task}, the obstacle can vary in size and position. The goal is to find policies that navigate the robot around the obstacle while completing the task as quickly as possible, without violating the safety constraints that require the end effector to maintain a safe distance from the obstacle.

We consider three task variations: 1) obstacle height, 2) obstacle width, and 3) obstacle position in a horizontal direction (left-right in Fig.~\ref{fig:obstacle-task}). The obstacle varies in height from \SI{0.049}{\meter}
 to \SI{0.331}{\meter} and in width from \SI{0.09}{\meter} to \SI{0.331}{\meter}. The horizontal position ranges from \SI{0.274}{\meter} to \SI{0.311}{\meter} with respect to the origin. We use Latin hypercube sampling to ensure a more even sample distribution and obtain 20 task variations from the specified ranges. We learn each variation for 120 iterations.

This learning problem formulation has three rewards: 1) a fixed success reward, 2) a goal distance reward, and 3) an obstacle avoidance reward. The fixed success reward assigns a fixed reward if the BT finishes successfully. The positive goal distance reward increases, the closer the end effector gets to the goal. The obstacle avoidance reward is a negative function that penalizes end-effector states that are close to the obstacle. These reward functions are combined to encourage fast execution while discouraging getting too close to the obstacle.
A policy is considered feasible if it satisfies two conditions: First, the end effector does not come closer to the obstacle than \SI{40}{\milli\meter}. Second, the policy must successfully complete the BT by bringing the end effector to the goal position.

\begin{figure}[tpb!!]
	{
		\setlength{\fboxrule}{0pt}
		\framebox{\parbox{3in}{
            \begin{center}
		      \includegraphics[width=0.95\columnwidth]{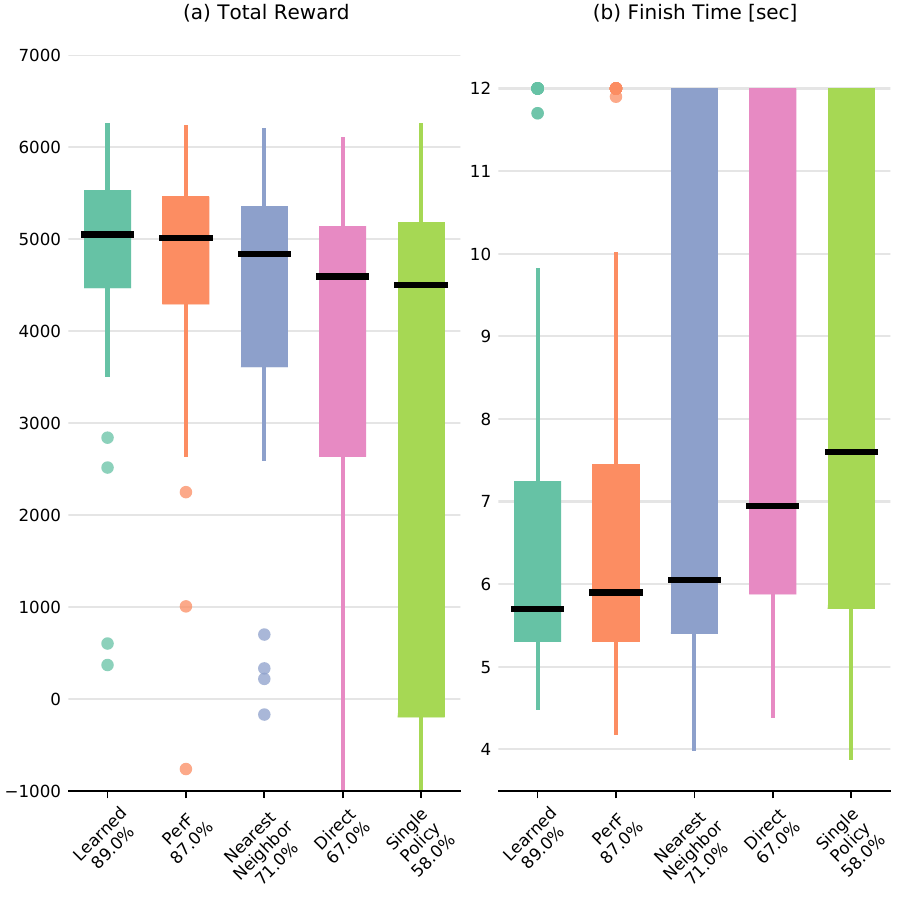}
                \includegraphics[width=0.85\columnwidth]{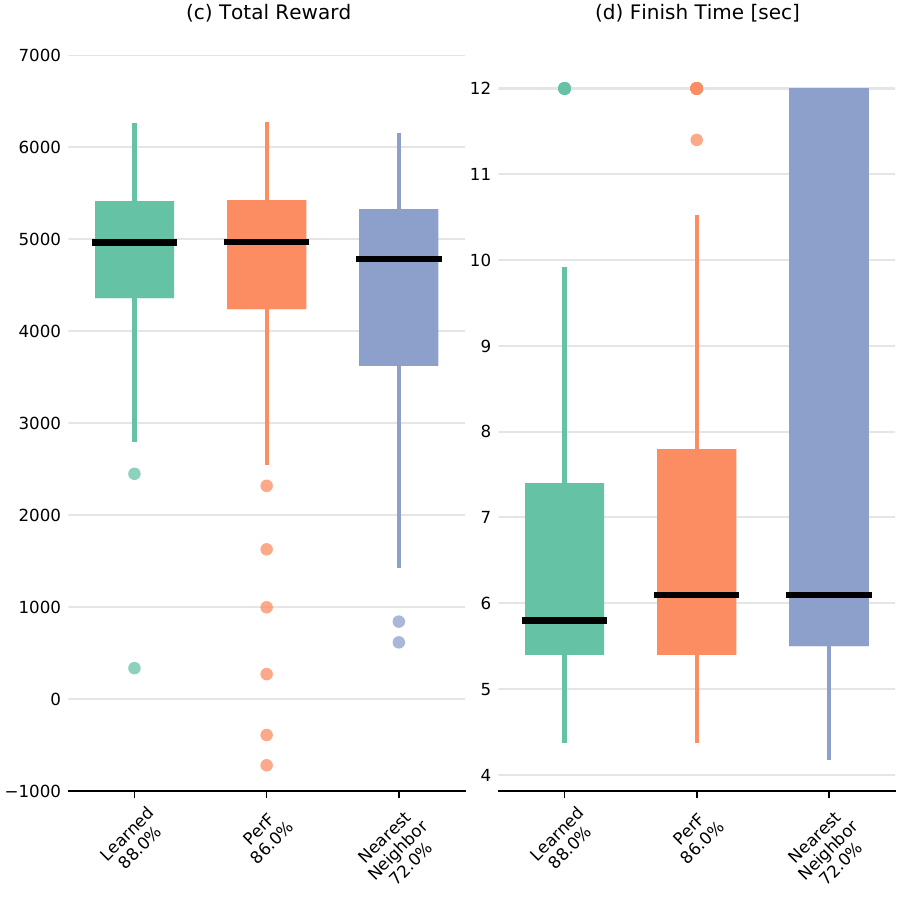}
            \end{center}
            }
	       }
    }
	\caption{The total reward (a, c) and the execution time (b, d) of the obstacle task in simulation (a, b) and on the real system (c, d). The box plots show the median (black line) and interquartile range ($25^{th}$ and $75^{th}$ percentile); the lines extend to the most extreme data points not considered outliers, and outliers are plotted individually. The success percentages are shown below the method names.}
	\label{fig:results-obstacle}
\end{figure}

The policy for this task has six learnable parameters consisting of two coordinates of the intermediate goal points and two thresholds to transition between goal points. A more detailed description of the task is provided in~\cite{mayr21iros, mayr2022skill}. Notably, the structure of this policy with its thresholds allows for different movement strategies. For example, for flat obstacles, the goal can be reached with only a single intermediate point, while larger obstacles require both intermediate points, as shown in Fig.~\ref{fig:obstacle-task}.
\subsubsection*{Results and Discussion}
For the evaluation, we randomly sample 20 new task variations ($\mathbb{V}_{test}$) that are not included in the training set, and compare the performance of our proposed model and the baseline methods. Specifically, we assess the execution time and the reward achieved by each parameter configuration in the new task variation. The reward value is chosen as a performance metric as it reflects how well a policy balances between the goal-reaching and obstacle-avoidance objectives expressed in the reward functions.

 The simulation results are shown in Fig.~\ref{fig:results-obstacle}a) and b) and Table~\ref{tab:results}. They show that the policies obtained by optimizing the output of our \textit{PerF} model performs similarly to the policies that are explicitly learned. Our model achieves a success percentage of 87\percent compared to the 89\percent of the learned ones and a total reward in a similar range. In contrast to that, the nearest neighbor baseline succeeds only in 71\percent of the variations. The \textit{direct} model also only achieves a success percentage of 67\percent and has significantly more outliers in the reward. Further investigation indicates that the reason for the low performance is that an interpolation between policies is often not valid. This is especially the case between motion configurations that use a single or both intermediate points.

Based on these results from simulation we also evaluated the learned policies, our model outputs and the nearest neighbor policies on the real robot system. Although this includes a transfer from simulation to the real system, the results shown in Fig.~\ref{fig:results-obstacle}c) and d) have only minor variations from the simulation results. This also demonstrates the robustness of this policy formulation as a whole.

\subsection{Push task}
\label{sec:exp-push}
The goal of this task is to push an object from a varying start location to a varying goal location. The object is shown in Fig.~\ref{fig:exp-setup} and has a skewed weight distribution with respect to its bounds.
 
We consider two types of task variations: 1) the starting position of the object in both horizontal directions and 2) the goal position of the object in both horizontal directions.
For the starting position, we consider samples from a circle with a diameter of \SI{0.16}{\meter} around a center point. For the goal position, a triangular-shaped region is used. Fig.~\ref{fig:exp-setup} shows the start and goal positions for a single repetition.

The learning formulation has two rewards: 1) the object position reward, which is a function of the difference between the actual and desired goal position, and 2) the object orientation reward, which is based on the difference between the actual and desired goal orientation. For our experiment, we prioritize the object position reward, which is weighted 10 times more heavily than the orientation reward.

Similarly to previous work~\cite{mayr2022combining, mayr2022skill}, the push task has four BTMG parameters that are learned. They are depicted in Fig.~\ref{fig:push-parameters}. These parameters control additional start and goal offsets in the horizontal directions $(x,y)$, determining the shape of the push vector that is indicated in Fig.~\ref{fig:push-parameters}. The start and goal orientation of the object for this task are fixed.

The object being pushed is an right-angled triangular object with dimensions \SI{0.3}{\meter} x \SI{0.15}{\meter} x \SI{0.07}{\meter}, and a weight of \SI{2.5}{\kilogram}. 
The tool on the end effector is a cubic peg with side lengths of \SI{45}{\milli\meter}
and therefore covers less than 15\percent of the side length of the object. 
In this task, the error between the desired goal position and orientation and the achieved one serves as direct performance measures for the policy.

\begin{figure}[tpb!!]
	{
		\setlength{\fboxrule}{0pt}
		\framebox{\parbox{3in}{
            \begin{center}
		      \includegraphics[width=0.95\columnwidth]{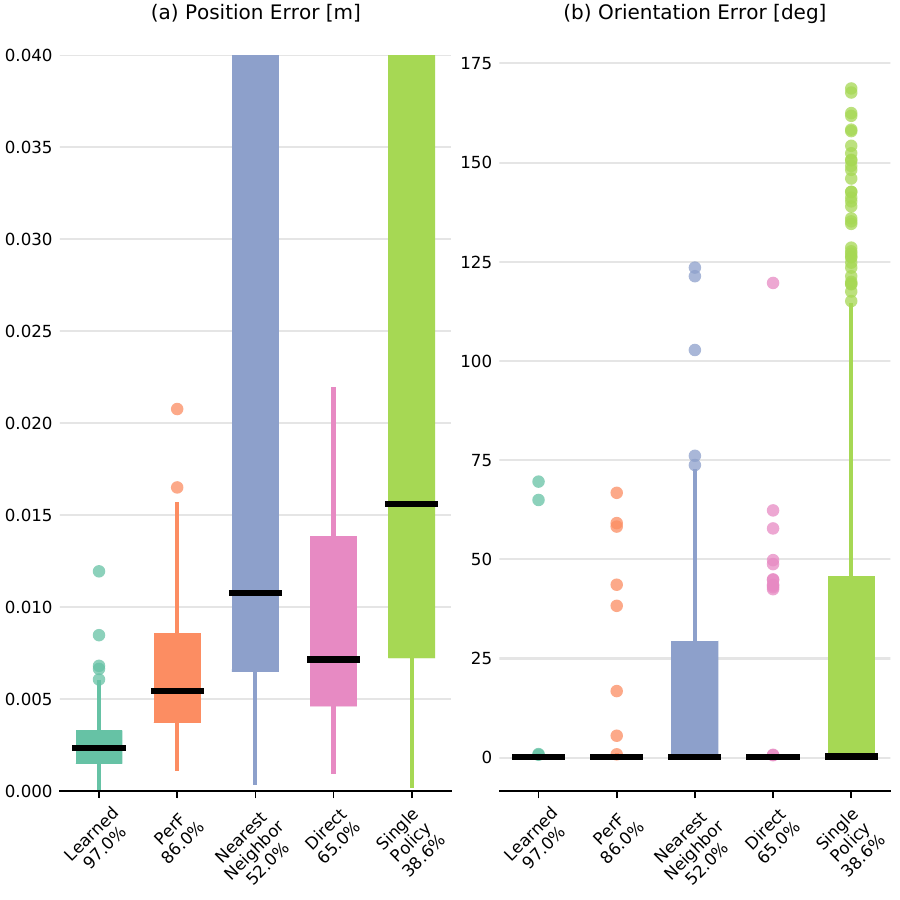}
                \includegraphics[width=0.85\columnwidth]{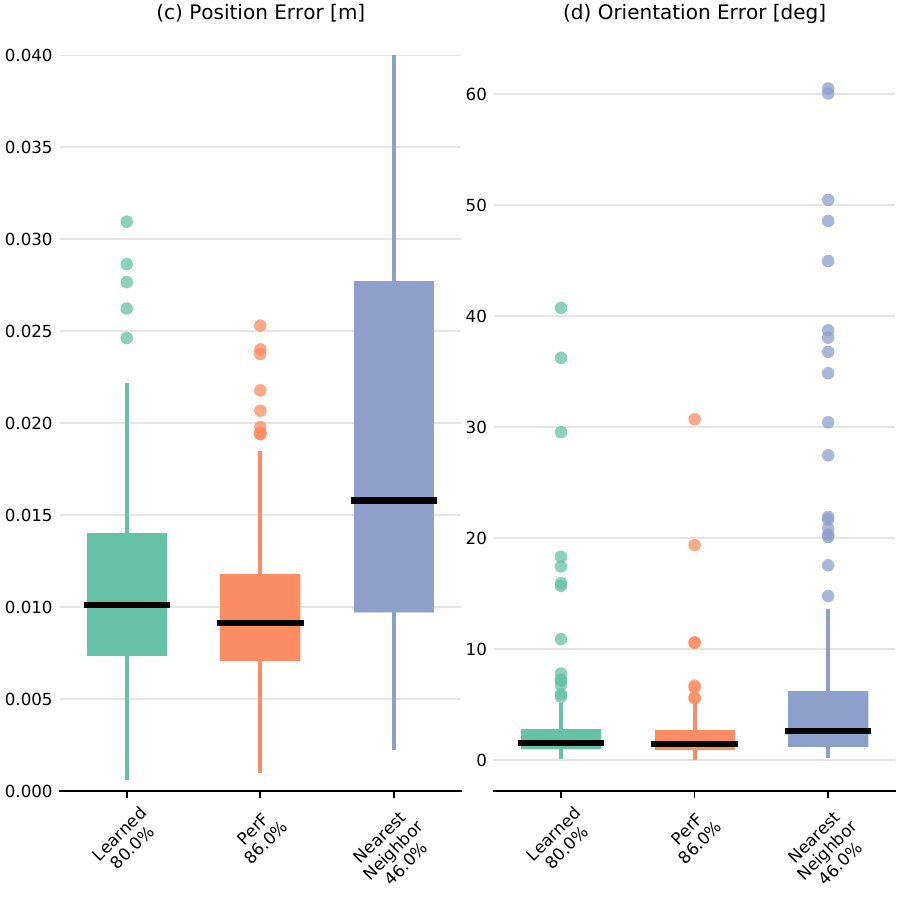}
            \end{center}
            }
            }
	}
	\caption{The final position error (a, c) and orientation error (b, d) of the push task in simulation (a, b) and on the real system (c, d).  The box plots show the median (black line) and interquartile range ($25^{th}$ and $75^{th}$ percentile); the lines extend to the most extreme data points not considered outliers, and outliers are plotted individually. The success percentages are shown below the method names.}
	\label{fig:results-push}
\end{figure}

\input{table2}

\subsubsection*{Results and Discussion}
The results for the simulation are shown in Figure \ref{fig:results-push}a) and b).  We consider a policy feasible if the position error between the goal location of the object and the desired goal location is less than \SI{11}{\milli\meter} and the orientation error is less than \SI{30}{\deg}.
The high success percentage of 97\percent for the learned policies shows that it is generally possible to solve this task. Our proposed model solves 86\percent of the configuration and outperforms all baselines that do not require explicit learning. The gap to the \textit{direct} model, which achieved a success rate of 65\percent, is significant. The nearest neighbor and the single policy approach only achieved 52\percent and 38\percent, which shows not only the difficulty of the task but also excludes them as practical solutions.

Similar to the obstacle task, we also executed the learned policies on the real robot system. To account for the differences of such a contact-rich task to the simulation, we increase the allowed final position error by \SI{4}{\milli\meter} but keep the same angular maximum.

The results for the evaluation on the real system are in Fig.~\ref{fig:results-push}c) and d) as well as in  Table~\ref{tab:results}. As intuitively expected, the success percentages generally drop as not all policies transfer to the real system. Similar to the evaluation in simulation, the nearest neighbor baseline performs poorly. However, it is notable that our model now outperforms the explicitly learned policies in both the success rate and the final error. A possible explanation for this is that our model needed to generalize, whereas an explicitly learned policy is able to exploit the simulation to the maximum extent possible. During the experiments, we also observed that policies from our model generally kept a larger distance from the object when approaching it and also had fewer collisions with it.

To determine the time efficiency of our approach, we compute time required to compute BTMG parameters for 60 new task variations. This analysis compares learning BTMG parameters from scratch using the RL-pipeline and obtaining BTMG parameters using our approach.
Starting from scratch with the RL-pipeline, median completion times were 770.315 seconds for the obstacle task and 1232.625 seconds for the push task. In contrast, the optimization phase of our approach achieved median completion times of 1.27 seconds for the obstacle task and 5.189 seconds for the push task. Additionally, obtaining a trained PERF model took an average of 66.628 seconds for the obstacle task and 317.025 seconds for the push task.
During optimization, we observed some outliers, likely stemming from the stochastic nature of the process. The analysis was performed on a laptop equipped with an Intel(R) Core(TM) i7-10870H CPU running at 2.20GHz with 8 physical cores and hyper-threading, along with 64GB of RAM.
\section{Conclusion and Future Work}
Agile robotics requires that a system adapts quickly to changing conditions. In this work, we introduced an extension to BTMGs, a motion representation based on behavior trees and motion generators, which addresses this challenge. Our approach enables the use of learned policies in previously unseen variations of a task, allowing for fast adaption of robot behavior to changes in the task or environment. 

The experimental evaluation demonstrates that our approach effectively learns a model capable of adapting to new task variations. Our method exhibits comparable performance to explicitly trained policies and consistently outperforms all other baseline models. Furthermore, experiments conducted on the real robotic system demonstrate the successful transferability of our approach from simulation to reality, even in a contact-rich task. Notably, our proposed method can even outperform explicitly learned policies in the same contact-rich task, indicating superior generalization capabilities.

In future work, it is worth exploring whether the uncertainty modeled by the GP can be leveraged to make more accurate predictions about successful execution. This uncertainty measure could also be used for out-of-distribution detection.
Another promising direction is to use the learned model to return policy parameters for task parameters, such as friction, for which the values are not known a priori. 
In this case, we could jointly optimize over both policy and task parameters to identify a compatible set of learned parameters. 

\bibliography{2023-IROS}
\bibliographystyle{bib/IEEEtran}

\end{document}

%% file: review.tex
\usepackage{amsmath,amssymb,amsfonts}
\usepackage{graphicx}
\usepackage{xcolor}
\usepackage{ulem}

\usepackage{ifthen}
\newboolean{showcomments}
\setboolean{showcomments}{true} 
\ifthenelse{\boolean{showcomments}}
  {\newcommand{\nb}[2]{
    \fcolorbox{gray}{yellow}{\bfseries\sffamily\scriptsize#1}
    {\sf\small$\blacktriangleright${#2}$\blacktriangleleft$}
  }
  
  }
  {\newcommand{\nb}[2]{}
  
  }

%% file: table2.tex
\begin{table*}[t]
\caption{The median performance values and the $25^{th}$ and $75^{th}$ percentiles for both tasks. A "-" indicates that configuration was not evaluated.}
\label{tab:results}
\resizebox{2\columnwidth}{!}{%
\begin{tabular}{@{}cccccccccccccccccc@{}}
\hline
\multirow{2}{*}{Task} &
  \multirow{2}{*}{\begin{tabular}[c]{@{}c@{}}Performance\\ Measure\end{tabular}} &
  \multirow{2}{*}{Environment} &
  \multicolumn{2}{|c}{\emph{Learned}} &
  \multicolumn{2}{|c}{\emph{PerF}} &
  \multicolumn{2}{|c}{\begin{tabular}[c]{@{}c@{}}\emph{Nearest}\\ \emph{Neighbor}\end{tabular}} &
  \multicolumn{2}{|c}{\emph{Direct}} &
  \multicolumn{2}{|c}{\emph{Single Policy}} \\ 
 &
   &
   &
  \multicolumn{1}{|c}{Median} &
  Percentiles &
  \multicolumn{1}{|c}{Median} &
  Percentiles &
  \multicolumn{1}{|c}{Median} &
  Percentiles &
  \multicolumn{1}{|c}{Median} &
  Percentiles &
  \multicolumn{1}{|c}{Median} &
  Percentiles \\ \hline
\multirow{4}{*}{Obstacle} &
  \multirow{2}{*}{Total Reward} &
  Simulation &
  \multicolumn{1}{|c}{5050} &
  (4467, 5531) &
  \multicolumn{1}{|c}{\textbf{5013}} &
  (4290, 5462) &
  \multicolumn{1}{|c}{4834} &
  (3607, 5357) &
  \multicolumn{1}{|c}{4594} &
  (2635, 5143) &
  \multicolumn{1}{|c}{4496} &
  (-200, 5184) \\ 
 &
   &
  Reality &
  \multicolumn{1}{|c}{4963} &
  (4357, 5414) &
  \multicolumn{1}{|c}{\textbf{4966}} &
  (4238, 5426) &
  \multicolumn{1}{|c}{4782} &
  (3625, 5327) &
  \multicolumn{1}{|c}{--} &
  -- &
  \multicolumn{1}{|c}{--} &
  -- \\ 
 &
  \multirow{2}{*}{\begin{tabular}[c]{@{}c@{}}Finish Time\\ {[}sec{]}\end{tabular}} &
  Simulation &
  \multicolumn{1}{|c}{5.7} &
  (5.3, 7.3) &
  \multicolumn{1}{|c}{\textbf{5.9}} &
  (5.3, 7.5) &
  \multicolumn{1}{|c}{6.1} &
  (5.4, 12) &
  \multicolumn{1}{|c}{7} &
  (5.9, 12) &
  \multicolumn{1}{|c}{7.6} &
  (5.7, 12) \\ 
 &
   &
  Reality &
  \multicolumn{1}{|c}{5.8} &
  (5.4, 7.4) &
  \multicolumn{1}{|c}{\textbf{6.1}} &
  (5.4, 7.8) &
  \multicolumn{1}{|c}{6.1} &
  (5.5, 12) &
  \multicolumn{1}{|c}{--} &
  -- &
  \multicolumn{1}{|c}{--} &
  -- \\ \hline
\multirow{4}{*}{Push} &
  \multirow{2}{*}{\begin{tabular}[c]{@{}c@{}}Position Error \\ {[}m{]}\end{tabular}} &
  Simulation &
  \multicolumn{1}{|c}{0.002} &
  (0.002, 0.003) &
  \multicolumn{1}{|c}{\textbf{0.006}} &
  (0.004, 0.009) &
  \multicolumn{1}{|c}{0.011} &
  (0.007, 0.083) &
  \multicolumn{1}{|c}{0.007} &
  (0.005, 0.014) &
  \multicolumn{1}{|c}{0.016} &
  (0.007, 0.123) \\ 
 &
   &
  Reality &
  \multicolumn{1}{|c}{0.01} &
  (0.007, 0.014) &
  \multicolumn{1}{|c}{\textbf{0.009}} &
  (0.007, 0.012) &
  \multicolumn{1}{|c}{0.016} &
  (0.01, 0.028) &
  \multicolumn{1}{|c}{--} &
  -- &
  \multicolumn{1}{|c}{--} &
  -- \\ 
 &
  \multirow{2}{*}{\begin{tabular}[c]{@{}c@{}}Orientation Error \\ {[}deg{]}\end{tabular}} &
  Simulation &
  \multicolumn{1}{|c}{0.15} &
  (0.07, 0.33) &
  \multicolumn{1}{|c}{\textbf{0.16}} &
  (0.07, 0.34) &
  \multicolumn{1}{|c}{0.18} &
  (0.08, 29.48) &
  \multicolumn{1}{|c}{0.11} &
  (0.06, 0.26) &
  \multicolumn{1}{|c}{0.29} &
  (0.1, 45.76) \\ 
 &
   &
  Reality &
  \multicolumn{1}{|c}{1.51} &
  (0.98, 2.76) &
  \multicolumn{1}{|c}{\textbf{1.42}} &
  (0.94, 2.74) &
  \multicolumn{1}{|c}{2.58} &
  (1.18, 6.23) &
  \multicolumn{1}{|c}{--} &
  -- &
  \multicolumn{1}{|c}{--} &
  -- \\ \hline
\end{tabular}%
}
\end{table*}